%% file: 1684.tex
\documentclass[runningheads]{llncs}
\usepackage{graphicx}
\usepackage{algorithm}
\usepackage{algpseudocode} 
\usepackage{amsmath,amssymb} 
\usepackage{color}

\input{commands.tex}

\begin{document}
\title{DYAN: A Dynamical Atoms-Based Network \\
 For Video Prediction\thanks{This work was supported in part by NSF grants IIS--1318145, ECCS--1404163, and CMMI--1638234; AFOSR grant FA9550-15-1-0392; and the Alert DHS Center of
Excellence under Award Number 2013-ST-061-ED0001.}} 

\titlerunning{DYAN}
%
\author{Wenqian Liu\orcidID{0000-0003-4274-8538} \and Abhishek Sharma\orcidID{0000-0001-6128-5124} \and  Octavia Camps\orcidID{0000-0003-1945-9172} \and Mario Sznaier\orcidID{0000-0003-4439-3988}}
%
\authorrunning{ W. Liu, A. Sharma, O. Camps, M. Sznaier}
%

\institute{Electrical and Computer Engineering, Northeastern University, Boston, MA 02115
\email{liu.wenqi,sharma.abhis@husky.neu.edu,  camps,msznaier@northeastern.edu  }\\
\url{http://robustsystems.coe.neu.edu} }
\maketitle              
\begin{abstract}
The ability to anticipate the future is essential when  making real time critical decisions,  provides valuable information to  understand dynamic natural scenes, and can help unsupervised video representation learning.  State-of-art video prediction is  based on complex architectures  that  need to learn large numbers of parameters, are potentially hard to train, slow to run, and may produce blurry predictions. In this paper, we introduce DYAN, a novel network with very few parameters and easy to train, which   produces accurate, high quality frame predictions,  faster than previous approaches.  DYAN owes its good qualities to  its encoder and decoder, which are designed following concepts from systems identification theory and  exploit the dynamics-based invariants of the data.  Extensive experiments using several standard video datasets show that DYAN is superior generating frames and that it generalizes well across domains.
\end{abstract}

\section{Introduction}

\begin{figure}[h]
\centering
\begin{tabular}{ccc}
\includegraphics[height=0.2\textwidth,width = 0.48\textwidth]{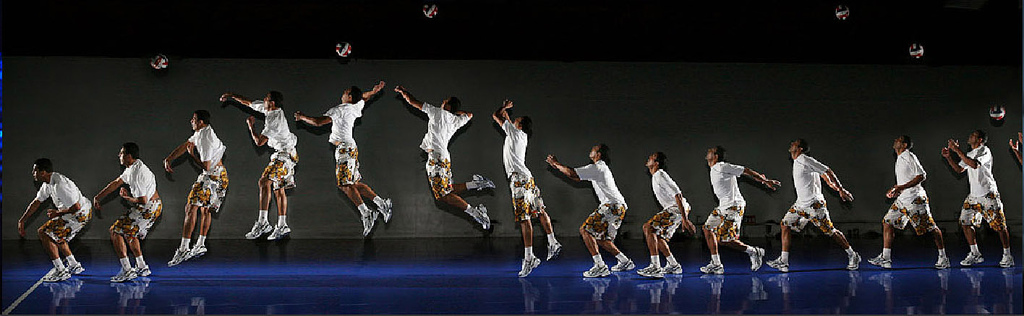} & \ &
\includegraphics[height=0.2\textwidth,width=0.48\textwidth]{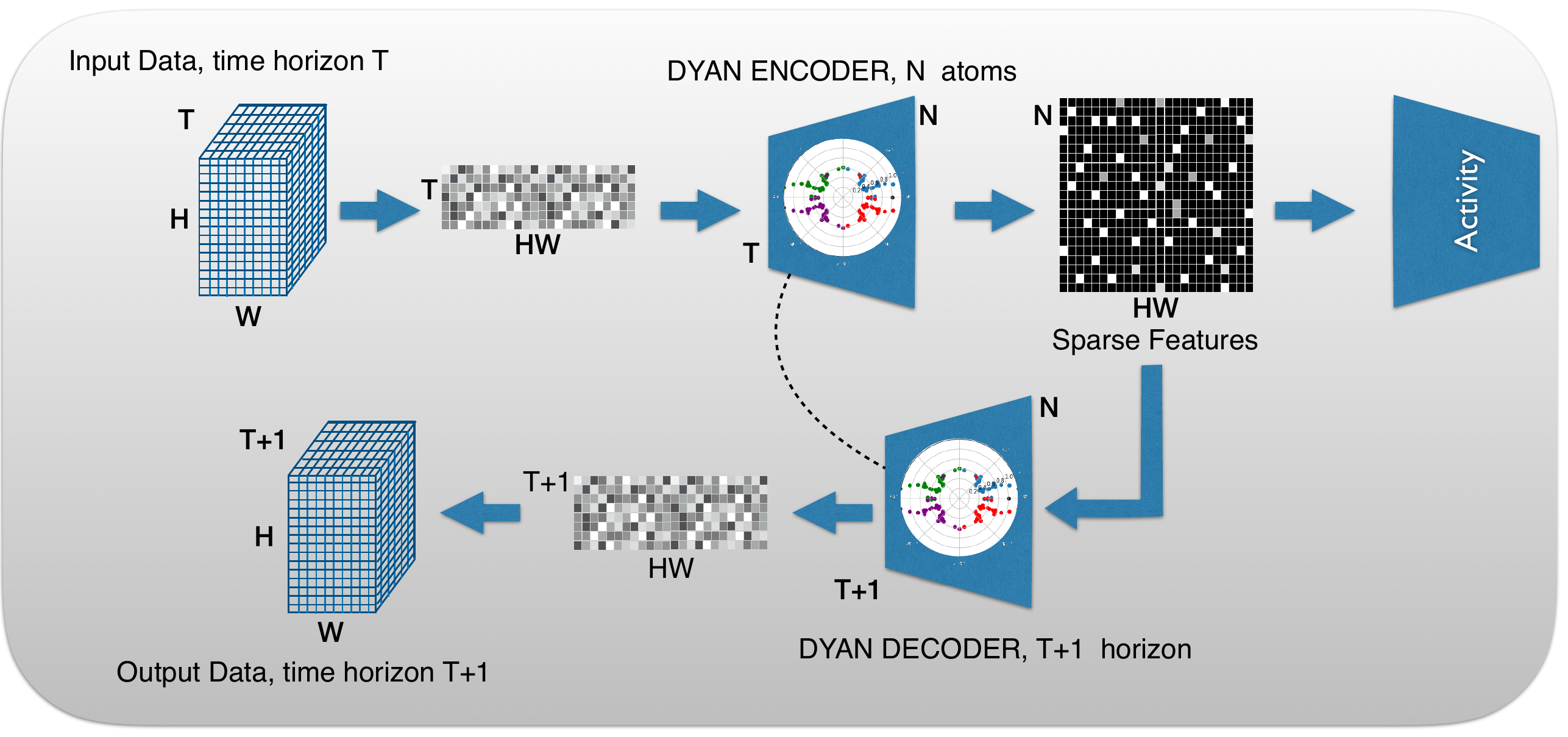}\\
(a) & \ & (b)
\end{tabular}
\caption{(a) Dynamics and motion provide powerful cues to understand scenes and predict the future. (b) DYAN's architecture: Given $T$ consecutive $H \times W$ frames, the network uses a dynamical atoms-based encoder to generate a set of sparse $N \times HW$ features that capture the  dynamics of each pixel, with $N \gg T$. These features can be passed to its dynamical atoms-based decoder to reconstruct the given frames and predict the next one, or they can be used for other tasks such as action classification.}\label{fig:motivation}
\end{figure}

The recent exponential growth in data collection capabilities and  the use of supervised deep learning approaches have helped to make tremendous progress in computer vision. However, learning good representations for the analysis and understanding of dynamic scenes, with limited or no supervision, remains a challenging task. This is in no small part due to the complexity of the changes in appearance and of the  motions that are observed in  video sequences of natural scenes.  
Yet, these changes and motions  provide powerful cues to understand dynamic scenes such as the one shown in  Figure~\ref{fig:motivation}(a), and they can be used to predict what is going to happen next. 
Furthermore, the ability of anticipating the future is  essential  to make decisions and take action in critical real time systems such as autonomous driving. Indeed, recent approaches to video understanding  \cite{liang2017dual,mathieu2015deep,srivastava2015unsupervised} suggest that being capable to accurately generate/predict  future frames in video sequences can help to  learn useful features with limited or no supervision. 

Predicting future frames to anticipate what is going to happen next  requires good generative models that can make forecasts based on the available past data. 
Recurrent Neural Networks (RNN) and in particular Long Short-Term Memory (LSTM) have  been widely used to process sequential data and make such predictions. Unfortunately, RNNs are hard to train due to the exploding and vanishing gradient problems. As a result, they can easily learn short term  but not long-term dependencies. On the other hand, LSTMs and the related Gated Recurrent Units (GRU), addressed the vanishing gradient problem and are easier to use. However, their design is ad-hoc, with  many components whose purpose is not easy to interpret \cite{jozefowicz2015empirical}.

More  recent approaches \cite{mathieu2015deep,zhou2016learning,xue2016visual,luc2017predicting}  advocate using generative adversarial network (GAN) learning \cite{goodfellow2014generative}. Intuitively, this is motivated by reasoning that the better the generative models, the better the prediction will be, and vice-versa: by learning how to distinguish predictions from real data, the network will learn better models. However, GANs are also reportedly hard to train, since training requires finding a Nash equilibrium of a game, which might be hard to get using gradient descent techniques.

In this paper, we present a novel DYnamical Atoms-based Network, DYAN, shown in Figure~\ref{fig:motivation}(b). DYAN is similar in spirit to LSTMs, in the sense that it  also captures short and long term dependencies. However, DYAN is designed using concepts from dynamic systems identification theory, which help to drastically reduce its size and provide easy interpretation of its parameters. By adopting ideas from atom-based system identification, DYAN learns a structured dictionary of atoms to exploit dynamics-based affine invariants in video data sequences. Using this dictionary, the network is able to capture actionable information from the dynamics of the data and map it into a set of very sparse features, which  can then be used  in video processing tasks, such as frame prediction, activity recognition, semantic segmentation, etc. 
We demonstrate the power of DYAN's autoencoding by using it to  generate future frames in video sequences. Our extensive experiments using several standard video datasets show that DYAN can predict future frames more accurately and  efficiently than current state-of-art approaches. 

In summary, the main contributions of this paper are:
\begin{itemize}
\item A novel auto-encoder network  that captures long and short term temporal information  and  explicitly incorporates  dynamics-based affine invariants;
\item The proposed network is  shallow, with very few parameters. It is easy to train and it does not take large disk space to save the learned model. 
\item The proposed network is  easy to interpret and it is easy to visualize what it learns, since the parameters of the network have a clear physical meaning.
\item The proposed network can predict future frames accurately and efficiently without introducing blurriness.
\item The model is differentiable, so it can be fine-tuned for another task if necessary. For example, the front end (encoder) of the proposed network can be easily incorporated at the front of other networks designed for video tasks such as activity recognition, semantic video segmentation, etc.
\end{itemize}

The rest of the paper is organized as follows. Section~\ref{sec:work} discusses related previous work. Section~\ref{sec:background} gives a brief summary of the concepts and procedures from dynamic systems theory, which are used in the design of DYAN.  Section~\ref{sec:approach} describes  the design of DYAN, its components and how it is trained. Section~\ref{sec:details} gives more details of the actual implementation of DYAN, followed by section~\ref{sec:experiments} where we report experiments comparing its performance in frame prediction against the state-of-art approaches. Finally, section~\ref{sec:conclusion} provides concluding remarks and directions for future applications of DYAN.

\section{Related Work} \label{sec:work}

There exist an extensive literature  devoted to the problem of extracting  optical flow from images \cite{horn1981determining}, including recent deep learning approaches \cite{dosovitskiy2015flownet,ilg2017flownet}. Most of these methods focus on  {\em Lagrangian} optical flow, where the flow field represents the displacement between corresponding pixels or features across frames. In contrast, DYAN can also work with {\em Eulerian} optical flow, where the motion is captured by the changes at individual pixels, without requiring finding correspondences or tracking features. Eulerian flow  has been shown to be useful for tasks such as motion enhancement \cite{wadhwa2013phase} and video frame interpolation \cite{meyer2015phase}.

State-of-art algorithms for action detection and recognition also exploit temporal information. Most deep learning approaches to action recognition use spatio-temporal data, starting with detections at the frame level \cite{saha2016deep,peng2016multi} and linking them across time by using very short-term temporal features such as optical flow. However, using such a short horizon misses the longer term dynamics of the action and can negatively impact performance.  This issue is often addressed by following up with some costly hierarchical aggregation over time.  More recently, some approaches detect tubelets  \cite{kalogeiton2017action,hou2017tube} starting with a longer temporal support than optical flow. However, they still rely on a relatively small number of frames, which is fixed a priori, regardless of the complexity of the action. Finally, most of these approaches  do not provide explicit encoding and decoding of the involved dynamics, which if available could be useful for inference and generative problems.

In contrast to the large volume of literature on action recognition and motion detection, there  are relatively few  approaches to frame prediction. 
Recurrent Neural Networks (RNN) and in particular Long Short-Term Memory (LSTM) have  been  used to predict frames. Ranzato et al. \cite{ranzato2014video} proposed a RNN to predict frames based on a discrete set of patch clusters, where an average of 64 overlapping tile predictions were used to avoid blockiness effects. In \cite{srivastava2015unsupervised} Srivastava et al. used instead an LSTM architecture with an ${\ell}_2$ loss function.   Both of these approaches produce blurry predictions due to using averaging. Other LSTM-based approaches include the work of Luo et al. \cite{luo2017unsupervised} using an encoding/decoding architecture with optical flow and the work of Kalchbrenner et al. \cite{kalchbrenner2016video} that estimates the probability distribution of the pixels. 

In \cite{mathieu2015deep}, Mathieu et al. used generative adversarial network (GAN) \cite{goodfellow2014generative} learning
 together with a multi-scale approach and a new loss based on image gradients to improve image sharpness in the predictions. Zhou and Berg \cite{zhou2016learning}  used a similar approach to predict future state of objects and Xue et al. \cite{xue2016visual} used a variational autoencoder to predict future frames from a single frame. More recently,  Luc et al. \cite{luc2017predicting} proposed an autoregressive convolutional network to predict semantic segmentations in future frames bypassing pixel prediction. Liu et al. \cite{liu2017video} introduced a network that synthesizes  frames by estimating voxel flow. However, it assumes that the optical flow is constant across multiple frames.  Finally, Liang et al. \cite{liang2017dual} proposed a dual motion GAN architecture that combines frame and flow predictions to generate future frames. All of these approaches involve large networks, potentially hard to train.
 
Lastly,  DYAN's encoder was inspired by the sparsification layers  introduced by Sun et al. in \cite{sun2017supervised} to perform image classification. However, DYAN's encoder is fundamentally different since it must use a {\em structured} dictionary  (see  (\ref{eq:complexdictionary})) in order to model dynamic data, while the sparsification layers in  \cite{sun2017supervised} do not. 

 \section{Background}\label{sec:background}

\subsection{Dynamics-based Invariants}

The power of {\em geometric} invariants  in  computer vision  has been recognized for a long time \cite{mundy1992geometric}. On the other hand,  {\em dynamics}-based affine invariants   have been used far less. These dynamics-based invariants, which were originally proposed for tracking \cite{ayazoglu2011dynamic}, activity recognition \cite{li2012cross}, and chronological sorting of images \cite{dicle2016solving}, tap on the properties of linear time invariant (LTI) dynamical systems. 
As briefly summarized below,
the main idea behind these invariants, is that if the available sequential data (i.e. the trajectory of a target being tracked or the values of a pixel as a function of time) can be  modeled as the output of some unknown LTI system, then, this underlying system has several attributes/properties that are invariant to affine transformations (i.e. viewpoint or illumination changes).  In this paper, as described in detail in section~\ref{sec:approach}, we  propose to use this affine invariance property to reduce the number of parameters in the proposed network, by leveraging the fact that multiple observations of one motion, captured in different conditions, can be described using one single set of these invariants.  

Let  $\cal S$  be a LTI system, described either by an autoregressive model or a state space model: 
\begin{eqnarray}
y_k  &=&  \sum_{i=1}^{n} a_iy_{k-i} \quad  \qquad
\ \ \ \text{ \% Autoregressive Representation} \label{eq:a} \\
\mat{x}_{k+1}&=&\mat{A}\mat{x}_k;  \;  y_k = \mat{C}\mat{x}_k   \qquad \text{\% State Space Representation} \label{eq:ss}  \\
 \text{with} \; 
\mat{x}_k&=&\begin{bmatrix}
y_{k-n} \\ \vdots \\y_k 
\end{bmatrix}, \; 
\mat{A}=\begin{bmatrix} 
0 & 1& \ldots &0 \\
\vdots & \ddots & \ddots & 0 \\
0 & 0& \ldots & 1 \\
a_n & a_{n-1} & \ldots & a_1 
\end{bmatrix}; \; 
\mat{C}=\begin{bmatrix} 0 &\ldots &0 &1 \end{bmatrix} \nonumber 
\end{eqnarray}
where  ${{y}}_k$\footnote{For simplicity of notation, we consider here $y_k$ scalar, but the invariants also hold for $\vect{y}_k \in \mathbb{R}^d$.} 
is the observation at time $k$, and $n$ is  the (unknown a priori) order of the model (memory of the system). Consider now a given initial condition $\mat{x}_o$ and its corresponding sequence $\mat{x}$. 
The Z-transform of a sequence $\mat{x}$ is defined as $X(z) = \sum_{k=0}^\infty x_k z^{-k}$, where $z$ is a complex variable $z = r e^{j\phi}$. Taking $Z$ transforms on both sides of \eqref{eq:ss} yields:
\beq \label{eq:gc} z (\mat{X}(z)-\mat{x}_o)=\mat{A}\mat{X}(z) \; \Rightarrow  \; \mat{X}(z)=z(z\mat{I}-\mat{A})^{-1}\mat{x}_o, \; Y(z)=z\mat{C}(z\mat{I}-\mat{A})^{-1}\mat{x}_o 
\eeq
where ${\cal G}(z) \doteq z\mat{C}(z\mat{I}-\mat{A})^{-1}$ is the transfer function from initial conditions to outputs. Using the explicit expression for the matrix inversion and assuming non-repeated poles, leads to
\beq \label{eq:impulse}
Y(z)=\frac{z\mat{C}_{\text{adj}}(z\mat{I}-\mat{A})\mat{x}_o}{\text{det}(z\mat{I}-\mat{A})}\doteq \sum_{i=1}^n \frac{z c_i}{z-p_i} \; \iff y_k = \sum_{i=1}^n c_i p_i^{k}, \; k=0,1,\ldots 
\eeq
where the roots of the denominator, $p_i$, are the eigenvalues of $\mat{A}$ (e.g. poles of the system) and the coefficients $c_i$ depend on the initial conditions.
Consider now an affine transformation $\Pi$. Then, substituting\footnote{(using homogeneous coordinates)} in
(\ref{eq:a}) we have,
$
y'_k \doteq\Pi ( y_k)  =  \Pi (\sum_{i=1}^{n} a_i{y}_{k-i}) = \sum_{i=1}^{n} a_i \Pi (y_{k-i}) 
$.
Hence, the order $n$, the model coefficients $a_i$ (and hence the poles $p_i$) are affine invariant since the sequence $y'_k$ is explained by the same autoregressive  model as the sequence $y_k$.

\subsection{LTI System Identification using Atoms}

Next, we briefly summarize an  atoms-based algorithm \cite{bekiroglu2018randomized} to  identify an LTI system from a given output sequence.

First, consider a set with an  infinite number of atoms, where each atom is the impulse response of a  LTI first order (or second order) system with a single real pole $p$ (or two conjugate complex poles, $p$ and $p^*$).  Their transfer functions  can be written as:
\[
{\cal{G}}_p(z) = \frac{w z}{z - p} \ \text{and} \ {\cal{G}}_p(z) = \frac{w z}{z - p} + \frac{w^* z}{z - p^*}
\]
where  $w \in \mathbb{C}$, and their impulse responses are given by $\mat{g}_p = w[1, p, p^2, p^3, \dots ]'$ and $\mat{g}_p = w[1, p, p^2, p^3, \dots ]' + w^* [1, p^*, {p^*}^2, {p^*}^3, \dots ]'$, for first and second order systems, respectively.

Next, from (\ref{eq:gc}),  every  proper transfer function  can be approximated to arbitrary precision as a linear combination of the above transfer functions\footnote{Provided that if a complex pole $p_i$ is used, then its conjugate $p_i^*$  is also used.}:
\[
{\cal{G}}(z) = \sum_{i} c_i {\cal{G}}_{p_i}(z)
\]
 Hence,  low order dynamical models can be estimated from output data $\mat{y} = [y_1, y_2,y_3,y_4, \dots]'$ by solving the following sparsification  problem:
\[
\min_{\mat{c} = \{c_i\}} \|\mat{c}\|_o \ \ 
\text{subject to: } \|\mat{y} -  \sum{c}_i\mat{g}_p\|_2^2 \le \eta^2
\]
where  $\|.\|_o$ denotes cardinality and the constraint imposes fidelity to the data.
Finally, note that solving the above optimization is not trivial  since minimizing cardinality is an NP-hard problem and the number of poles to consider is infinite.  The authors in \cite{bekiroglu2018randomized} proposed to address these issues by 1) using the ${\ell}_1$ norm relaxation for cardinality,  2) using  impulse responses of the atoms truncated to the length of  the available data, and 3) using a finite set of atoms with uniformly sampled poles in the unit disk. Then, using these ideas one could solve instead:

\beq
\min_{\mat{c}} \frac{1}{2}\|\mat{y}_{1:T} - {D^{(T)} \mat{c}}\|_2^2 + \lambda \|\mat{c}\|_1 \label{eq:elasticnet}
\eeq
where $\mat{y}_{1:T} = [y_1, y_2, \dots, y_T]'$, ${D^{(T)}}$ is a {\em structured} dictionary matrix with $T$ rows and $N$ columns:
 \beq
{D}^{(T)} = \left [ \begin{array}{cccc}
 p_1^0 & p_2^0 & \dots & p_N^0 \\
 p_1 & p_2 & \dots & p_N \\
 p_1^2 & p_2^2 & \dots & p_N^2 \\
 \vdots& \vdots & \vdots & \vdots \\
 p_1^{T-1} & p_2^{T-1} & \dots & p_N^{T-1}\\
 \end{array}
 \right ] 
 \label{eq:complexdictionary}
 \eeq
where each column  corresponds to the impulse response of a pole $p_i$, $i=1,\dots,N$ inside or near the unit disk in $\mathbb{C}$. Note that the dictionary is completely parameterized by the magnitude and phase of its poles.

\section{DYAN: A dynamical atoms-based network}\label{sec:approach}

\begin{figure}[h]

\centering
\includegraphics[width=0.85\textwidth]{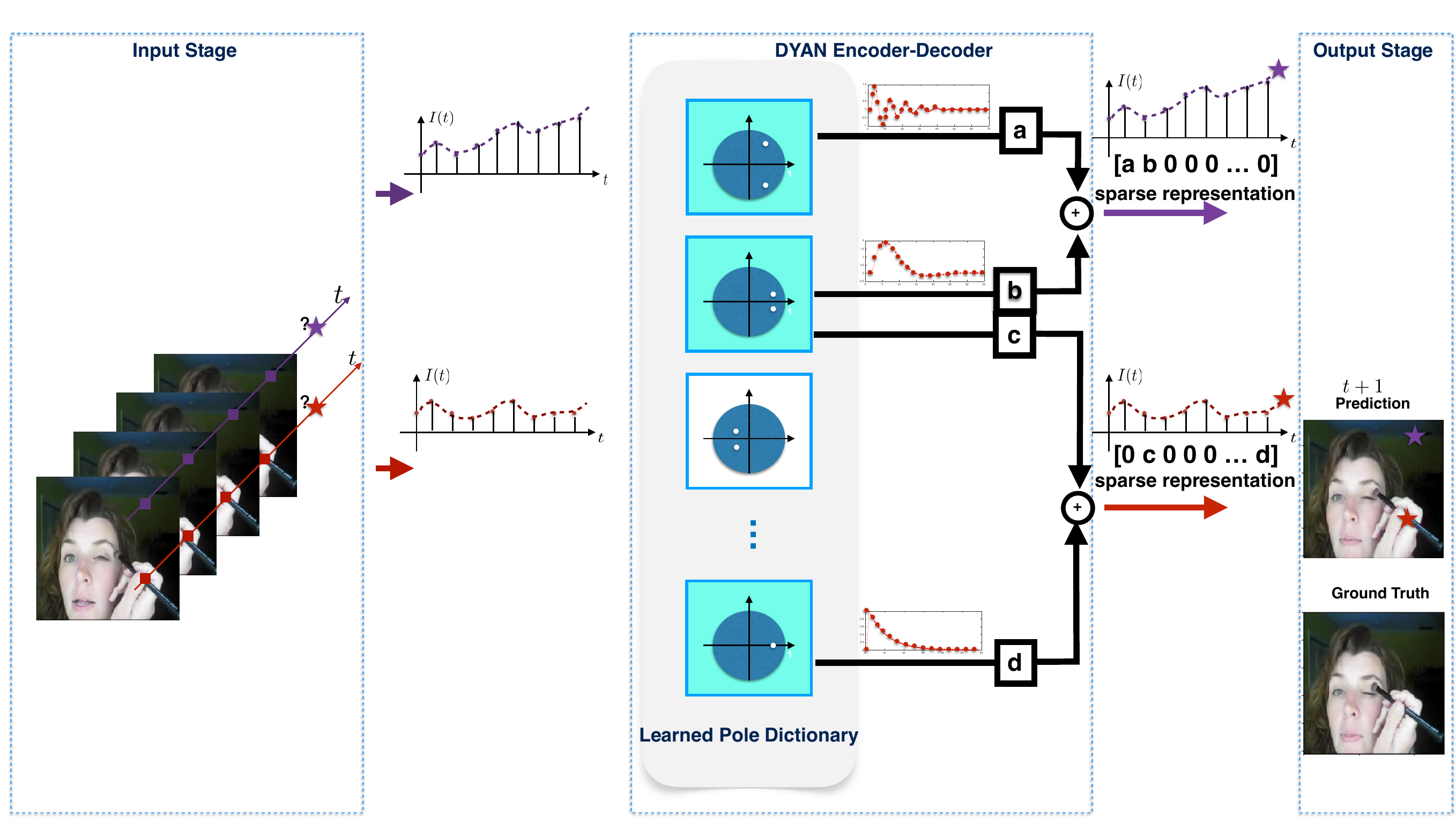}
\caption{  DYAN  identifies the dynamics for each pixel,  expressing them as a linear combination of  a small subset of dynamics-based atoms from a   dictionary (learned during training).  The selected atoms  and the corresponding coefficients are represented using  sparse feature vectors, found by a sparsification step.  These features  are  used by the decoder to reconstruct the input data and predict the next frame by using the same dictionary, but with an extended temporal horizon. See text for more details.}
\label{fig:pipe}
\end{figure}

In this section we describe in detail the  architecture of {\bf DYAN}, a dynamical atoms-based network. Figure~\ref{fig:motivation}(b) shows its block diagram, depicting its two main components: a dynamics-based encoder and dynamics-based decoder. Figure ~\ref{fig:pipe} illustrates how these two modules work together to capture the dynamics at each pixel,  reconstruct the input data and predict future frames.

The goal of DYAN is to capture the dynamics of the input by mapping them  to a  latent space, which is learned during training, and to provide the inverse mapping from this feature space back to the input domain.  The implicit assumption is that the dynamics of the input data should have a sparse representation in this latent  space, and that this representation should be enough to reconstruct the input and to predict future frames. 

Following the ideas from dynamic system identification presented in section~\ref{sec:background}, we propose to use as  latent space, the space spanned by a set of atoms that are  the impulse responses of a set of first (single real pole) and second order (pair of  complex conjugate poles) LTI systems, as illustrated in Figure~\ref{fig:pipe}. However, instead of using a set of random poles in the unit disk as proposed in \cite{bekiroglu2018randomized}, the proposed network learns a  set of ``good'' poles by minimizing a loss function that penalizes  reconstruction and predictive poor quality.

The main advantages of the DYAN architecture are:
\begin{itemize}
\item {\bf Compactness:} Each pole in the dictionary can be used by  more than one pixel, and  affine invariance  allows to re-use the same poles, even if the data was captured under different conditions from the ones used in training. Thus, the total number of poles needed to have a rich dictionary, capable of modeling the dynamics of a wide range of inputs, is relatively small. Our experiments show that the total number of parameters of the dictionary, which are the magnitude and phase of its poles, can be below  two hundred and the network still produces high quality frame predictions.

\item{\bf Adaptiveness to the dynamics complexity:} The network  adapts to the complexity of the  dynamics of the input by automatically deciding how many atoms it needs to use to explain them. The more complex the dynamics, the higher the order of the model is needed, i.e. the higher the number of  atoms will be selected, and the longer term memory of the data will be used by the decoder to  reconstruct and predict  frames.

\item {\bf Interpretable:} Similarly to CNNs that learn sets of convolutional filters, which can be easily visualized, DYAN learns a basis of very simple dynamic systems, which are also easy to visualize by looking at their poles and impulse responses.

\item {\bf Performance:}   
Since pixels are processed in parallel, independently of each other\footnote{On the other hand, if modeling cross-pixel correlations is desired, it is easy to modify the network to process jointly local neighborhoods using a group Lasso optimization in the encoder.}, blurring in the predicted frames  and computational time are both reduced.

\end{itemize}

\subsection{DYAN's encoder}

The encoder stage takes as input a set of $T$ consecutive $H\times W$ frames (or features), which are flattened into   $HW$, $T\times 1$  vectors, as shown in Figure~\ref{fig:motivation}(b). Let one of these vectors be  $\mat{y}_l$. Then, the output of the encoder is the collection of the minimizers of $HM$ sparsification  optimization problems:
\beq
\mat{c}_l^* = \arg\min_{\mat{c}} \frac{1}{2} \|\mat{y}_l - {D}^{(T)}\mat{c} \|^2_2 + \lambda \|\mat{c}\|_1 \qquad l = 1, \dots,HW
\label{eq:elasticnet2}
\eeq
where ${D}^{(T)}$ is the dictionary with the learned atoms, which is shared by all pixels and $\lambda$ is a regularization parameter.
Thus, using a  $T \times N$ dictionary, the output of the encoder stage is a set of sparse $HW$ $N \times 1$ vectors, that can be reshaped into  $H\times W \times N$ features.

In order to avoid working with complex poles $p_i$,  we  use instead a dictionary $D^{(T)}_{\rho,\psi}$ with columns corresponding to the real and imaginary parts of increasing powers of the poles $p_i = \rho_i e^{j\psi_i}$ in the first quadrant ($0\le \psi_i\le \pi/2$), of their conjugates and of their mirror images in the third and fourth quadrant\footnote{But eliminating duplicate columns.}:  $\rho_i^k \cos(k\psi_i)$,  $\rho_i^k \sin(k\psi_i)$, $(-\rho_i)^k \cos(k\psi_i)$, and $(-\rho_i)^k \sin(k\psi_i)$ with $k=0,\dots,T-1$. In addition, we include a fixed atom at $p_i = 1$ to model constant inputs.
\beq
 {D^{(T)}_{\rho,\psi}} = \left [ \begin{array}{ccccc}
 1 &1 & 0 & \dots &0  \\
 1 &\rho_1\cos\psi_1 &  \rho_1\sin\psi_1   & \dots & -\rho_N\sin\psi_N  \\
1 & \rho_1^2 \cos2\psi_1&  \rho_1^2 \sin2\psi_1& \dots & (-\rho_N)^2\sin2\psi_N \\
 \vdots & \vdots& \vdots &\vdots  & \vdots   \\
 1 &\rho_1^{T-1}\cos(T-1)\psi_1 & \rho_1^{T-1}\sin(T-1)\psi_1  & \dots
 &(-\rho_N)^{T-1}\sin(T-1)\psi_N\\
 \end{array}
 \right ]
 \eeq
Note that while equation (\ref{eq:elasticnet}) finds one $\mat{c}^*$ (and a set of poles) for each feature $\mat{y}$, it is trivial to process all the features in parallel with significant computational time savings. Furthermore, (\ref{eq:elasticnet}) can be easily  modified to force neighboring features, or features at the same location but from different channels, to select the same  poles by using a group Lasso formulation.

\begin{algorithm}
\caption{FISTA}\label{fista}
\begin{algorithmic}[1]
\Require Dictionary ${D} \in \mathbb{R}^{n\times m}$, input signal ${y} \in \mathbb{R}^n$, $\lambda$, $L$ the largest eigenvalue of $D^T D$, $A=I - \frac{1}{L}(D^T D)$, $b = \frac{1}{L} D^T y$ , $g = \frac{1}{L}$. Initialize iterator $t=0$, $c_t = 0 \in \mathbb{R}^m$, $\gamma_t = 0 \in\mathbb{R}^m$, $s_0 = 1$.
\While{stopping criterion not satisfied}
\State $\gamma = Ac_t  + b$
\State if $\gamma > g : c_{t+1} \leftarrow \gamma - g $ 
\State else $ \gamma < -g : c_{t+1} \leftarrow \gamma + g $
\State $s_{t+1}\leftarrow (1+\sqrt{(1+4s_t^2)})/2$
\State $ c_{t} \leftarrow c_{t+1}((s_t-1)/s_{t+1} + 1)) - c_{t}((s_t-1)/s_{t+1})$
\State $t \leftarrow t+1$
\EndWhile\\

\Return sparse code $c_t$
\end{algorithmic}
\end{algorithm}

In principle, there are available several sparse recovery algorithms that could be used to solve Problem (\ref{eq:elasticnet2}), including  LARS \cite{hesterberg2008least}, ISTA and FISTA\cite{beck2009fast}, and LISTA \cite{gregor2010learning}. Unfortunately, the structure of the dictionary needed here does not admit a matrix factorization of its Gram kernel, making the LISTA algorithm a poor choice in this case \cite{moreau2017understanding}. Thus, we chose to use FISTA, shown in Algorithm \ref{fista}, since  very efficient GPU implementations of this algorithm are available.

\subsection{DYAN's decoder}

The decoder stage takes as input the output of the encoder, i.e. a set of sparse $HW$ $N\times 1$ vectors and multiplies them with the encoder dictionary, extended with one more row: 
\beq
  \left [ \begin{array}{ccccc}
  1 &\rho_1^{T}\cos(T\psi_1) & \rho_1^{T}\sin(T\psi_1)  & \dots
 &(-\rho_N)^{T}\sin(T\psi_N)\\
 \end{array}
 \right ]
 \eeq
 to reconstruct the $T$ input frames and to predict the $T+1$ frame. Thus, the output of the decoder is a set of $HW$ $(T+1)\times 1$ vectors that can be reshaped into $(T+1)$, $H \times W$ frames.
 
\subsection{DYAN's training}

The parameters of the dictionary are learned using Steepest Gradient Descent (SGD) and the $\ell_2$ loss function. The back propagation rules for  the {\em encoder, decoder} layers can be derived by taking the subgradient of the empirical loss function with respect to the magnitudes and phases of the first quadrant poles  and the regularizing parameters. Here, for simplicity, we give the derivation for  $D^{(T)}_p$, but the one for $D^{(T)}_{\rho,\psi}$ can be derived in a similar manner.

Let $\mat{c}^*$ be the solution of one of the minimization problems in (\ref{eq:elasticnet}), where we dropped the subscript $l$ and the superscript  $(T)$ to simplify notation, and define
 \[
 {\cal F} = \frac{1}{2}\|\mat{y} - {D \mat{c}^*}\|_2^2 + \lambda \sum_{i=1}^N c_i^* \text{sign}(c_i^*)
 \]
Taking subgradients with respect to $\mat{c}^*$:
\[
\frac{\partial{\cal F}}{\partial \mat{c}^*} = 0 = -D^T(\mat{y} - D\mat{c}^*) + \lambda \vect{v} = 0
\]
where $\mat{v} = \left [ \begin{array}{ccc}v_1 &  \dots & v_N\end{array}\right ]^T$,  $v_i = $ sign$(c_i^*)$ if $c_i^*\neq 0$, and $v_i =  g$, where $-1 \le g \le 1$, otherwise. Then,
\[
\mat{c}^* = (D_\Lambda^TD_\Lambda)^{-1}\left [D^T_\Lambda\mat{y} - \lambda\mat{v}\right ]
\]
and
\[
\left .\frac{\partial{{\mat{c}^*}}}{\partial{D_{ij}}}\right |_\Lambda = 
(D^T_\Lambda D_\Lambda)^{-1} \left [ \frac{\partial{D^T_\Lambda\mat{y}}}{\partial{D_{ij}}}  - \frac{\partial{D^T_\Lambda D_\Lambda}}{\partial{D_{ij}}}{\mat{c}^*}\right]
\]
where the subscript $\left . . \right|_\Lambda$ denotes the active set of the sparse code $\mat{c}$, $D_\Lambda$ is composed of the active columns of $D$, and $\mat{c}_\Lambda$ is the vector with the active elements of the sparse code. Using the structure of the dictionary, we have
\[
\frac{\partial{\mat{c}^*_\Lambda}}{\partial{p_k}} =
\sum_{i=1}^M (i-1)p_k^{i-2} \frac{\partial{\mat{c}^*_\Lambda}}{\partial{D_{ik}}}; \ 
\frac{\partial{\mat{c}^*_\Lambda}}{\partial{y_j}}=
(D^T_\Lambda D_\Lambda )^{-1} \frac{\partial{D^T_\Lambda\mat{y}}}{\partial{y_j}};\ 
\frac{\partial{\mat{c}^*_\Lambda}}{\partial{\lambda}} = 
-(D^T_\Lambda D_\Lambda )^{-1}\text{sign}(\mat{c}^*_\Lambda)
\]
\begin{figure}[h]
\centering
\includegraphics[width=1\textwidth]{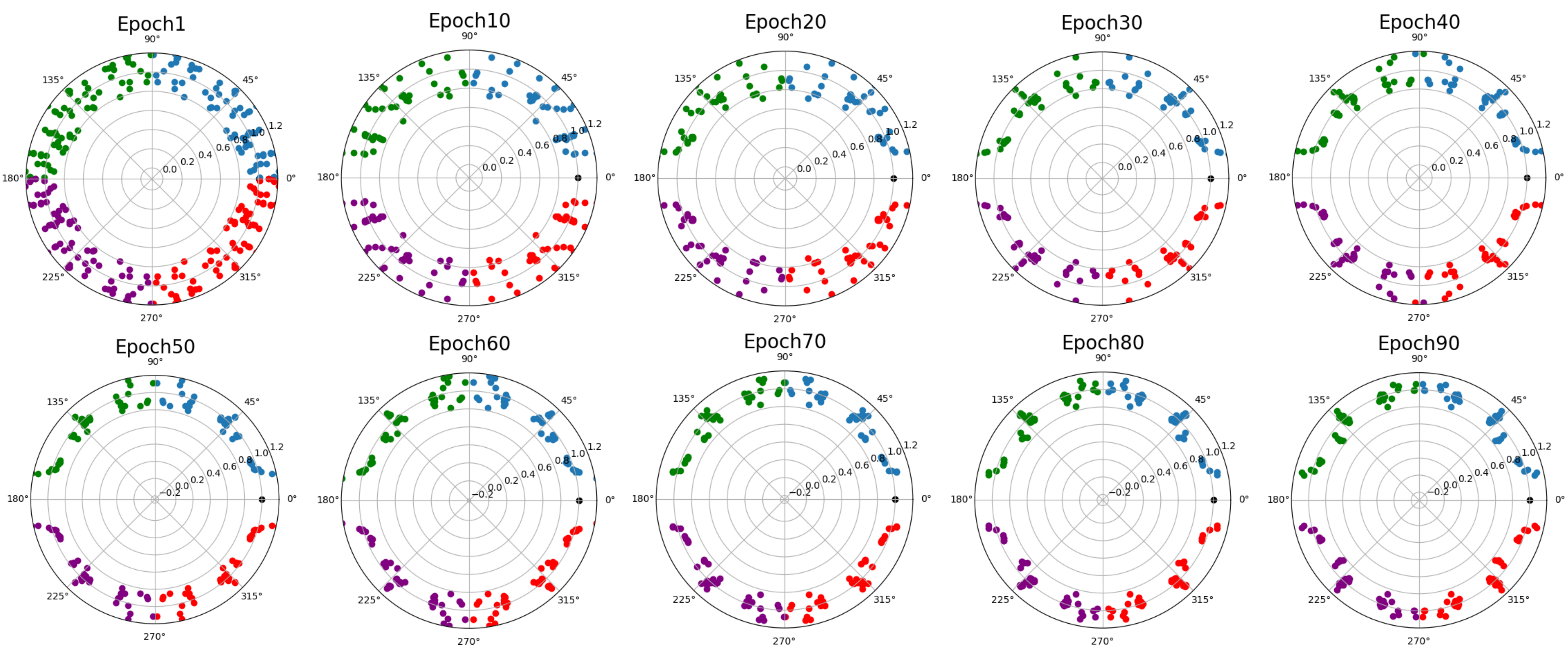}
\caption{Temporal evolution of a  dictionary  trained with the KITTI  dataset. }
\label{fig:poles}
\end{figure}

Figure~\ref{fig:poles} shows how a set of 160 uniformly distributed  poles within a ring around the unit circle move while training DYAN with videos from the KITTI video dataset \cite{geiger2013vision}, using the above back propagation and a $\ell_2$ loss function.  As shown in the figure, after only 1 epoch, the poles have already moved significantly and after 30 epochs the poles move slower and slower.

\section{Implementation Details}\label{sec:details}

We implemented\footnote{Code will be made available in Github.} DYAN using Pytorch version-0.3. A DYAN trained using  raw pixels as input produces nearly perfect reconstruction of the input frames. However, predicted frames may exhibit small lags at edges due to changes in pixel visibility. This problem can be easily addressed  by training DYAN using optical flow as input. {Therefore, given a video with $F$ input frames, we use coarse to fine optical flow \cite{pathakCVPR17learning} to obtain $T = F-1$ optical flow frames.} Then, we use these optical flow frames to predict with DYAN the next optical flow frame  to warp  frame $F$ into the predicted frame $F+1$. The dictionary is initialized with $40$ poles, uniformly distributed on a grid of $0.05 \times 0.05$ in the first quadrant within 
a ring around the unit circle defined by $0.85 \leq \rho \leq 1.15$,  their 3 mirror images in the other quadrants, and a fixed pole at $p=1$. Hence, the resulting  encoder  and decoder dictionaries have $N = 161$ columns\footnote{Note that the dictionaries do not have repeated columns, for example conjugate poles share the column corresponding to their real parts, so the number of columns is equal to the number of poles.} and $T$ and $T+1$ rows, respectively.   Each of the columns in the encoding dictionary was normalized to have norm 1. The maximum number of iterations for the FISTA step was set to 100.

\section{Experiments}\label{sec:experiments}
In this section, we describe a set of experiments using DYAN to predict the next frame and compare its performance against the state-of-art video  prediction algorithms. The experiments were run on  widely used public datasets, and illustrate the generative and generalization capabilities of our network.

\begin{figure}[th]
\centering
\includegraphics[width=1\textwidth]{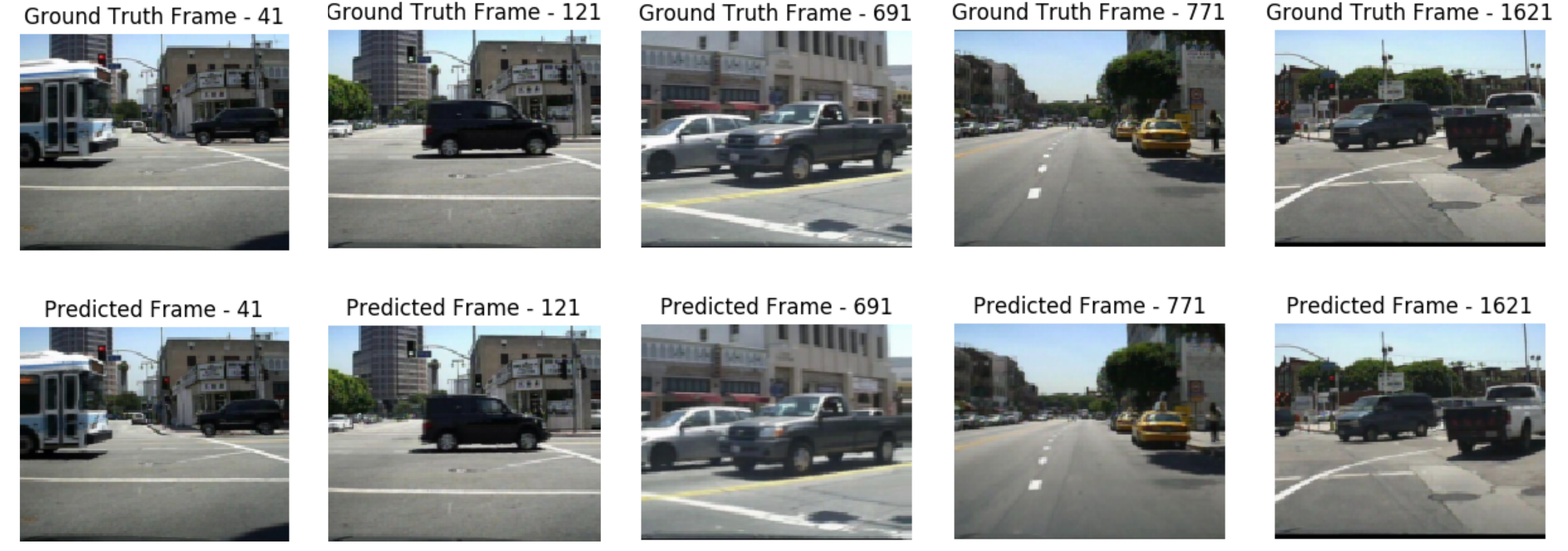}
\caption{Qualitative results for our model trained on the KITTI dataset and tested on the Caltech dataset, without fine tuning. The figure shows  examples from Caltech test set S10, sequence V010, with ground truth on the top row and predicted frames below.   As shown in the figure, our model produces sharp images and fully captures the motion of the vehicles and the camera.}
\label{fig:caltech}
\end{figure}

\subsection{Car Mounted Camera Videos Dataset}
We first evaluate our model on street view videos taken by car mounted cameras. Following the experiments settings in \cite{liang2017dual}, we trained our model on the KITTI dataset \cite{geiger2013vision}, including 57 recoding sessions (around 41k frames), from the City, Residential, and Road categories. Frames were center-cropped and resized to $128\times160$ as done in   \cite{lotter2016deep}. For these experiments, we trained our model with 10 input frames ($F=10, T=9$) and $\lambda = 0.01$ to predict frame 11. Then, we directly tested our model {\em without fine tuning} on the Caltech Pedestrian dataset \cite{dollar2009pedestrian}, testing partition (4 sets of videos), which consists of 66 video sequences. During testing time, each sequence was split into sequences of 10 frames, and frames were also center-cropped and resized to $128\times 160$. Also following \cite{liang2017dual},  the quality of the predictions for these experiments was measured using  MSE\cite{lotter2016deep}  and SSIM\cite{wang2004image}  scores, where  lower  MSE and  higher SSIM indicate better prediction results.

Qualitative results on the Caltech dataset are shown in Figure~\ref{fig:caltech}, where it can be seen that our model accurately predicts sharp, future frames. Also note that even though in this sequence there are cars moving towards  opposite directions or occluding each other, our model can  predict all motions well.
We compared DYAN's performance against three  state-of-the-art approaches: DualMoGAN\cite{liang2017dual}, BeyondMSE\cite{mathieu2015deep} and Prednet\cite{lotter2016deep}. For a fair comparison, we normalized our image values between 0 and 1 before computing the MSE score. As shown in Table \ref{caltech}, our model outperforms all other algorithms, even without fine tuning on the new dataset. This result shows the superior predictive ability of DYAN, as well as its transferability. 

For these experiments, the network was trained on 2 NVIDIA TITAN XP GPUs, using one GPU for each of the  optical flow channels. The model was trained for 200 epochs and it only takes 3KB to store it on  disk. Training only takes 10 seconds/epoch, and it takes an average of 230ms (including warping) to predict the next frame, given a sequence of 10 input frames. In comparison,   \cite{liang2017dual} takes  300ms to predict a frame.  

\begin{table}[]
\centering
\caption{MSE and SSIM scores of next frame prediction test on Caltech dataset after training on KITTI datset.}
\label{caltech}
\begin{tabular}{|c|c|c|c|c|c|}
\hline
Caltech &\begin{tabular}[c]{@{}c@{}}CopyLast\\(F=10)\end{tabular} & \begin{tabular}[c]{@{}c@{}}BeyondMSE \cite{mathieu2015deep} \\ (F = 10)\end{tabular} & \begin{tabular}[c]{@{}c@{}}PredNet \cite{lotter2016deep}\\  (F = 10)\end{tabular}  & \begin{tabular}[c]{@{}c@{}}DualMoGan \cite{liang2017dual} \\ (F = 10)\end{tabular} & \begin{tabular}[c]{@{}c@{}}Ours\\  (F = 10)\end{tabular} \\ \hline
MSE   &0.00795  & 0.00326                                               & 0.00313 & 0.00241                                              & {\color{red}\textbf{0.00087}}                                         \\ \hline
SSIM   &0.762 & 0.881                                                 & 0.884   & 0.899                                                & {\color{red}\textbf{0.952}}                                            \\ \hline
\end{tabular}
\end{table}

\begin{figure}[t]
\centering
\includegraphics[width=1\textwidth]{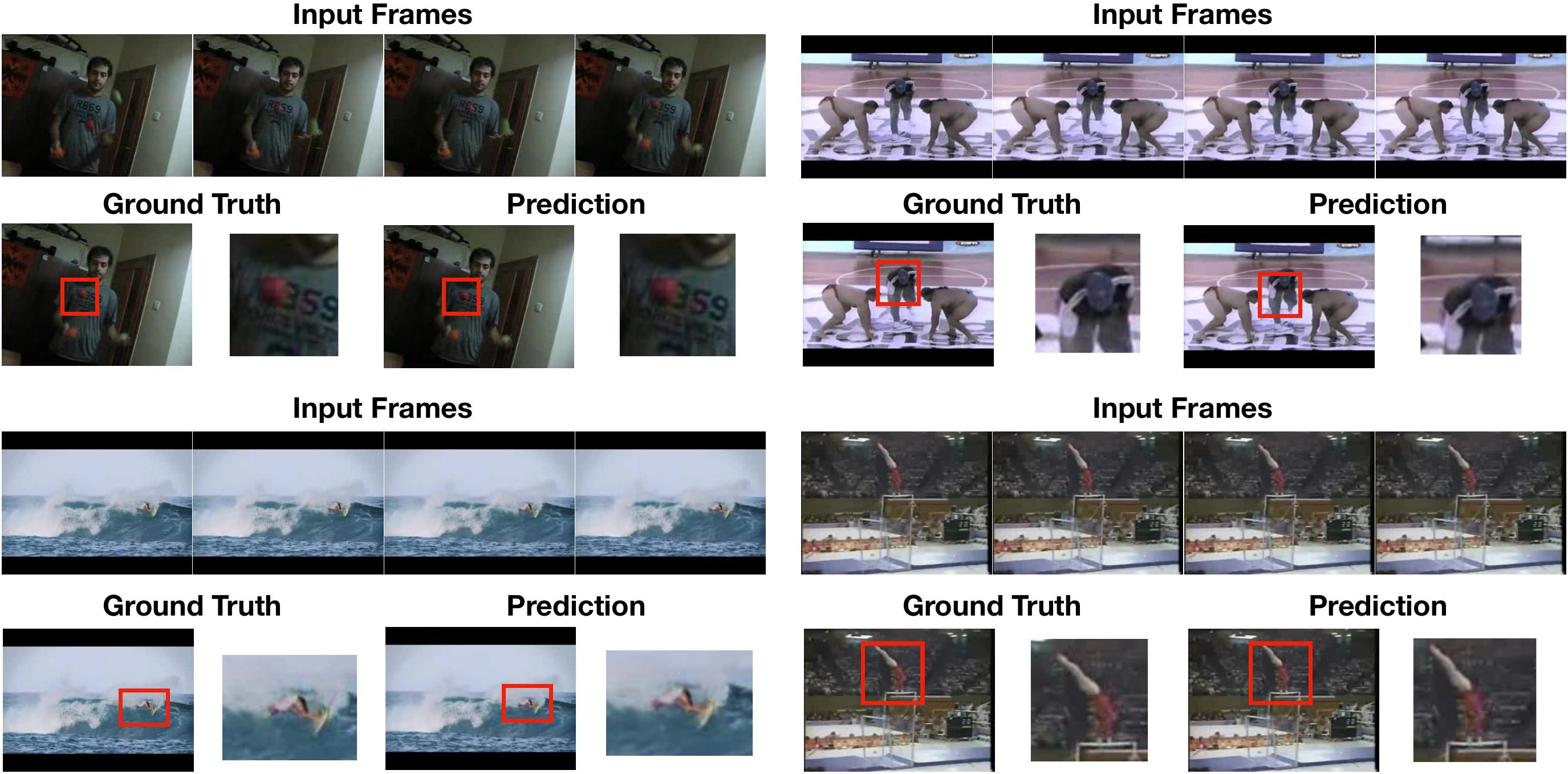}
\caption{Qualitative results for next frame prediction test on UCF-101. For each sequence, the first row shows the 4 input frames,  while the ground truth and our prediction are shown on the second row. We also enlarge the main moving portion inside each frame to show how similar our predictions are compared to the ground truth.}
\label{result1}
\end{figure}

\subsection{Human Action Videos Dataset}
We also tested  DYAN on generic  videos from the UCF-101 dataset \cite{soomro2012ucf101}. This  dataset contains 13,320 videos under 101 different action categories with an average length of 6.2 seconds.  Input frames are $240 \times 320$.  Following state-of-art algorithms \cite{liu2017video} and \cite{mathieu2015deep}, we trained using the first split and  using $F= 4$ frames as  input to predict the 5th frame. While testing, we adopted the test set provided by \cite{mathieu2015deep} and the evaluation script and optical masks provided by \cite{liu2017video} to mask in only the moving object(s) within each frame, resized to $256 \times 256$. There are in total 378 video sequences in the test set: every 10th video sequence was extracted from UCF-101 test list and then  5 consecutive frames are used, 4 for input and 1 for ground truth. Quantitative results with  PSNR\cite{mathieu2015deep} and SSIM\cite{wang2004image} scores, where the higher the score the better the prediction, are given in Table~\ref{ucf} and qualitative results are shown in Figure~\ref{result1}.  These experiments show that DYAN predictions achieve superior PSNR and SSIM scores  by identifying the dynamics of the optical flow instead of assuming it is constant as DVF does.  

\begin{figure}[h]
\centering
\includegraphics[width=0.45\textwidth]{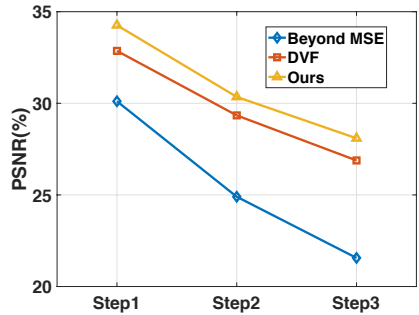}
\caption{Qualitative result for our model trained on UCF-101 dataset with $F = 4$. Other scores were obtained by running the code provided by the respective authors. All scores were computed using masks from \cite{liu2017video}.}
\label{fig:predicate}
\end{figure}

Finally, we also conducted a multi-step prediction experiment in which we applied our $F = 4$ model to  predict the next three future frames, where each prediction was used as a new available input frame. Figure~\ref{fig:predicate} shows the results of this experiment,   compared against the  scores for BeyondMSE \cite{mathieu2015deep} and DVF \cite{liu2017video}, where it can be seen that the PSNR scores of DYAN's predictions are consistently higher than the ones obtained using previous approaches.

For these experiments, DYAN was trained on 2 NVIDIA GeForce GTX GPUs, using one GPU for each of the  optical flow channels. Training takes around 65 minutes/epoch, and  predicting one frame takes 390ms (including warping).  Training converged at 7 epochs for $F=4$. In contrast, DVF takes severals day to train. DYAN's saved model only takes 3KB on hard disk.

\begin{table}[]
\centering
\caption{PSNR and SSIM scores of next frame prediction on UCF-101 dataset. Results for \cite{mathieu2015deep,liu2017video} were obtained by running the code provided by the respective authors.}
\label{ucf}
\resizebox{\linewidth}{!}{%
\begin{tabular}{|c|c|c|c|c|c|}
\hline
UCF-101 & \begin{tabular}[c]{@{}c@{}}CopyLast\\(F = 4)\end{tabular} & \begin{tabular}[c]{@{}c@{}}BeyondMSE \cite{mathieu2015deep} \\(F =4)\end{tabular} & \begin{tabular}[c]{@{}c@{}}OpticalFlow \cite{mathieu2015deep}\\(F = 4)\end{tabular} & \begin{tabular}[c]{@{}c@{}}DVF \cite{liu2017video}  \\(F = 4)\end{tabular}         & \begin{tabular}[c]{@{}c@{}}Ours \\ (F = 4)\end{tabular}  \\ \hline
PSNR   & 28.6   & 30.11                                                  & 31.6                                                    & 32.86          & {\color{red}\textbf{ 34.26}}                                             \\ \hline
SSIM   &0.89  & 0.88                                                  & 0.93                                                    & {0.93} & {\color{red}\textbf{0.96}}                                                                                                         \\ \hline
\end{tabular}
}
\end{table}

\section{Conclusion} \label{sec:conclusion}

We introduced a novel DYnamical Atoms-based Network, DYAN, designed using concepts from dynamic systems identification theory, to capture dynamics-based invariants in  video sequences, and to predict future frames. DYAN has several advantages compared to architectures previously used for similar tasks: it is compact,  easy   to train, visualize and interpret, it is fast to train, it produces high quality predictions fast, and generalizes well across domains. Finally, the high quality of DYAN's predictions  show that the sparse features  learned by its encoder  do capture the underlying dynamics of the input, suggesting that they will be useful for other unsupervised learning and video processing tasks such as activity recognition and video semantic segmentation.
\newpage
\bibliographystyle{splncs04}
\bibliography{1684}

\end{document}

%% file: commands.tex
\newcommand{\vect}[1]{\mathbf{#1}}
\newcommand{\mat}[1]{\mathbf{#1}}

\def\beq{\begin{equation}}
\def\eeq{\end{equation}}

%% file: 1684.bbl
\begin{thebibliography}{10}
\providecommand{\url}[1]{\texttt{#1}}
\providecommand{\urlprefix}{URL }
\providecommand{\doi}[1]{https://doi.org/#1}

\bibitem{ayazoglu2011dynamic}
Ayazoglu, M., Li, B., Dicle, C., Sznaier, M., Camps, O.I.: Dynamic
  subspace-based coordinated multicamera tracking. In: Computer Vision (ICCV),
  2011 IEEE International Conference on. pp. 2462--2469. IEEE (2011)

\bibitem{beck2009fast}
Beck, A., Teboulle, M.: A fast iterative shrinkage-thresholding algorithm for
  linear inverse problems. SIAM journal on imaging sciences  \textbf{2}(1),
  183--202 (2009)

\bibitem{dicle2016solving}
Dicle, C., Yilmaz, B., Camps, O., Sznaier, M.: Solving temporal puzzles. In:
  CVPR. pp. 5896--5905 (2016)

\bibitem{dollar2009pedestrian}
Doll{\'a}r, P., Wojek, C., Schiele, B., Perona, P.: Pedestrian detection: A
  benchmark. In: Computer Vision and Pattern Recognition, 2009. CVPR 2009. IEEE
  Conference on. pp. 304--311. IEEE (2009)

\bibitem{dosovitskiy2015flownet}
Dosovitskiy, A., Fischer, P., Ilg, E., Hausser, P., Hazirbas, C., Golkov, V.,
  van~der Smagt, P., Cremers, D., Brox, T.: Flownet: Learning optical flow with
  convolutional networks. In: Proceedings of the IEEE International Conference
  on Computer Vision. pp. 2758--2766 (2015)

\bibitem{geiger2013vision}
Geiger, A., Lenz, P., Stiller, C., Urtasun, R.: Vision meets robotics: The
  kitti dataset. The International Journal of Robotics Research
  \textbf{32}(11),  1231--1237 (2013)

\bibitem{goodfellow2014generative}
Goodfellow, I., Pouget-Abadie, J., Mirza, M., Xu, B., Warde-Farley, D., Ozair,
  S., Courville, A., Bengio, Y.: Generative adversarial nets. In: Advances in
  neural information processing systems. pp. 2672--2680 (2014)

\bibitem{gregor2010learning}
Gregor, K., LeCun, Y.: Learning fast approximations of sparse coding. In:
  Proceedings of the 27th International Conference on Machine Learning
  (ICML-10). pp. 399--406 (2010)

\bibitem{hesterberg2008least}
Hesterberg, T., Choi, N.H., Meier, L., Fraley, C.: Least angle and l1 penalized
  regression: A review. Statistics Surveys  \textbf{2},  61--93 (2008)

\bibitem{horn1981determining}
Horn, B.K., Schunck, B.G.: Determining optical flow. Artificial intelligence
  \textbf{17}(1-3),  185--203 (1981)

\bibitem{hou2017tube}
Hou, R., Chen, C., Shah, M.: Tube convolutional neural network (t-cnn) for
  action detection in videos. arXiv preprint arXiv:1703.10664  (2017)

\bibitem{ilg2017flownet}
Ilg, E., Mayer, N., Saikia, T., Keuper, M., Dosovitskiy, A., Brox, T.: Flownet
  2.0: Evolution of optical flow estimation with deep networks. In: IEEE
  Conference on Computer Vision and Pattern Recognition (CVPR). vol.~2 (2017)

\bibitem{jozefowicz2015empirical}
Jozefowicz, R., Zaremba, W., Sutskever, I.: An empirical exploration of
  recurrent network architectures. In: ICML. pp. 2342--2350 (2015)

\bibitem{kalchbrenner2016video}
Kalchbrenner, N., Oord, A.v.d., Simonyan, K., Danihelka, I., Vinyals, O.,
  Graves, A., Kavukcuoglu, K.: Video pixel networks. arXiv preprint
  arXiv:1610.00527  (2016)

\bibitem{kalogeiton2017action}
Kalogeiton, V., Weinzaepfel, P., Ferrari, V., Schmid, C.: Action tubelet
  detector for spatio-temporal action localization. arXiv preprint
  arXiv:1705.01861  (2017)

\bibitem{li2012cross}
Li, B., Camps, O.I., Sznaier, M.: Cross-view activity recognition using
  hankelets. In: Computer Vision and Pattern Recognition (CVPR), 2012 IEEE
  Conference on. pp. 1362--1369. IEEE (2012)

\bibitem{liang2017dual}
Liang, X., Lee, L., Dai, W., Xing, E.P.: Dual motion gan for future-flow
  embedded video prediction. arXiv preprint  (2017)

\bibitem{liu2017video}
Liu, Z., Yeh, R., Tang, X., Liu, Y., Agarwala, A.: Video frame synthesis using
  deep voxel flow. In: International Conference on Computer Vision (ICCV).
  vol.~2 (2017)

\bibitem{lotter2016deep}
Lotter, W., Kreiman, G., Cox, D.: Deep predictive coding networks for video
  prediction and unsupervised learning. arXiv preprint arXiv:1605.08104  (2016)

\bibitem{luc2017predicting}
Luc, P., Neverova, N., Couprie, C., Verbeek, J., LeCun, Y.: Predicting deeper
  into the future of semantic segmentation. In: of: ICCV 2017-International
  Conference on Computer Vision. p.~10 (2017)

\bibitem{luo2017unsupervised}
Luo, Z., Peng, B., Huang, D.A., Alahi, A., Fei-Fei, L.: Unsupervised learning
  of long-term motion dynamics for videos. arXiv preprint arXiv:1701.01821
  \textbf{2} (2017)

\bibitem{mathieu2015deep}
Mathieu, M., Couprie, C., LeCun, Y.: Deep multi-scale video prediction beyond
  mean square error. arXiv preprint arXiv:1511.05440  (2015)

\bibitem{meyer2015phase}
Meyer, S., Wang, O., Zimmer, H., Grosse, M., Sorkine-Hornung, A.: Phase-based
  frame interpolation for video. In: Proceedings of the IEEE Conference on
  Computer Vision and Pattern Recognition. pp. 1410--1418 (2015)

\bibitem{moreau2017understanding}
Moreau, T., Bruna, J.: Understanding the learned iterative soft thresholding
  algorithm with matrix factorization. arXiv preprint arXiv:1706.01338  (2017)

\bibitem{mundy1992geometric}
Mundy, J.L., Zisserman, A.: Geometric invariance in computer vision, vol.~92.
  MIT press Cambridge, MA (1992)

\bibitem{pathakCVPR17learning}
Pathak, D., Girshick, R., Doll\'{a}r, P., Darrell, T., Hariharan, B.: Learning
  features by watching objects move. In: Computer Vision and Pattern
  Recognition ({CVPR}) (2017)

\bibitem{peng2016multi}
Peng, X., Schmid, C.: Multi-region two-stream r-cnn for action detection. In:
  ECCV. pp. 744--759. Springer (2016)

\bibitem{ranzato2014video}
Ranzato, M., Szlam, A., Bruna, J., Mathieu, M., Collobert, R., Chopra, S.:
  Video (language) modeling: a baseline for generative models of natural
  videos. arXiv preprint arXiv:1412.6604  (2014)

\bibitem{saha2016deep}
Saha, S., Singh, G., Sapienza, M., Torr, P.H., Cuzzolin, F.: Deep learning for
  detecting multiple space-time action tubes in videos. arXiv preprint
  arXiv:1608.01529  (2016)

\bibitem{soomro2012ucf101}
Soomro, K., Zamir, A.R., Shah, M.: Ucf101: A dataset of 101 human actions
  classes from videos in the wild. arXiv preprint arXiv:1212.0402  (2012)

\bibitem{srivastava2015unsupervised}
Srivastava, N., Mansimov, E., Salakhudinov, R.: Unsupervised learning of video
  representations using lstms. In: International conference on machine
  learning. pp. 843--852 (2015)

\bibitem{sun2017supervised}
Sun, X., Nasrabadi, N.M., Tran, T.D.: Supervised multilayer sparse coding
  networks for image classification. arXiv preprint arXiv:1701.08349  (2017)

\bibitem{wadhwa2013phase}
Wadhwa, N., Rubinstein, M., Durand, F., Freeman, W.T.: Phase-based video motion
  processing. ACM Transactions on Graphics (TOG)  \textbf{32}(4), ~80 (2013)

\bibitem{wang2004image}
Wang, Z., Bovik, A.C., Sheikh, H.R., Simoncelli, E.P.: Image quality
  assessment: from error visibility to structural similarity. IEEE transactions
  on image processing  \textbf{13}(4),  600--612 (2004)

\bibitem{xue2016visual}
Xue, T., Wu, J., Bouman, K., Freeman, B.: Visual dynamics: Probabilistic future
  frame synthesis via cross convolutional networks. In: Advances in Neural
  Information Processing Systems. pp. 91--99 (2016)

\bibitem{bekiroglu2018randomized}
Yilmaz, B., Bekiroglu, K., Lagoa, C., Sznaier, M.: A randomized algorithm for
  parsimonious model identification. IEEE Transactions on Automatic Control
  \textbf{63}(2),  532--539 (2018)

\bibitem{zhou2016learning}
Zhou, Y., Berg, T.L.: Learning temporal transformations from time-lapse videos.
  In: European Conference on Computer Vision. pp. 262--277 (2016)

\end{thebibliography}
